\documentclass[10pt,twocolumn,letterpaper]{article}

\usepackage{cvpr}              
\usepackage[accsupp]{axessibility} 
\usepackage{cite}
\usepackage{comment}
\usepackage{amsmath,amssymb,amsfonts}
\usepackage{booktabs}       
\usepackage{adjustbox}
\usepackage{algorithm,algorithmic}
\usepackage{graphicx}
\usepackage{wrapfig}
\usepackage{textcomp}
\usepackage{url}            
\usepackage{amssymb}
\usepackage{multirow} 
\usepackage{soul} 
\usepackage{dsfont} 
\usepackage{indentfirst}  
\usepackage{natbib}

\def\ie{\emph{i.e.}}

\def\BibTeX{{\rm B\kern-.05em{\sc i\kern-.025em b}\kern-.08em
    T\kern-.1667em\lower.7ex\hbox{E}\kern-.125emX}}
\usepackage[dvipsnames]{xcolor}

\definecolor{cvprblue}{rgb}{0.21,0.49,0.74}
\definecolor{expgreen}{RGB}{8, 155, 100}

\usepackage[pagebackref,breaklinks,colorlinks,citecolor=cvprblue]{hyperref}

\def\paperID{11080} 
\def\confName{CVPR}
\def\confYear{2024}

\title{Cooperation Does Matter: Exploring Multi-Order Bilateral Relations for Audio-Visual Segmentation}   

\author{Qi Yang$^{1,2}$ \quad Xing Nie$^{1,2}$ \quad Tong Li$^{3}$ \quad Pengfei Gao$^{3}$ \quad Ying Guo$^{3}$ \\ 
Cheng Zhen$^{3}$ \quad Pengfei Yan$^{3}$ \quad Shiming Xiang$^{1,2}$ \\
${^1}~$School of Artificial Intelligence, University of Chinese Academy of Sciences~(UCAS) \quad \\ $^{2}~$Institute of Automation, Chinese Academy of Sciences~(CASIA) \quad $^{3}~$Meituan
}

\begin{document}
\maketitle   

\begin{abstract}
Recently, an audio-visual segmentation~(AVS) task has been introduced, aiming to group pixels with sounding objects within a given video. This task necessitates a first-ever audio-driven pixel-level understanding of the scene, posing significant challenges.
In this paper, we propose an innovative audio-visual transformer framework, termed COMBO, an acronym for COoperation of Multi-order Bilateral relatiOns. For the first time, our framework explores three types of bilateral entanglements within AVS: pixel entanglement, modality entanglement, and temporal entanglement. Regarding pixel entanglement, we employ a Siam-Encoder Module (SEM) that leverages prior knowledge to generate more precise visual features from the foundational model. For modality entanglement, we design a Bilateral-Fusion Module (BFM), enabling COMBO to align corresponding visual and auditory signals bi-directionally. As for temporal entanglement, we introduce an innovative adaptive inter-frame consistency loss according to the inherent rules of temporal. Comprehensive experiments and ablation studies on AVSBench-object~(84.7 mIoU on S4, 59.2 mIou on MS3) and AVSBench-semantic~(42.1 mIoU on AVSS) datasets demonstrate that COMBO surpasses previous state-of-the-art methods. Project page is available at \href{https://yannqi.github.io/AVS-COMBO}{https://yannqi.github.io/AVS-COMBO}.

\end{abstract}

\begin{figure}[t!]
    \centering
    \includegraphics[width=1.0\columnwidth]{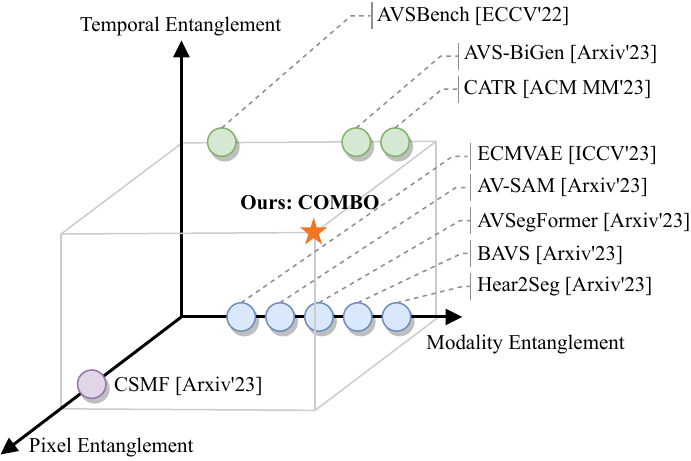}
    \vspace{-0.5cm}
    \caption{
    Comparison between the proposed COMBO and existing state-of-the-art methods. Our COMBO is the first work to simultaneously explore multi-order bilateral relations in modality, temporal and pixel levels.
    } 
    \label{fig:figure_task}
    \vspace{-0.5cm}
\end{figure}

\section{Introduction}

Human visual attention is often \emph{hear-guided}, \ie, we tend to focus on the object with sounds \cite{2020vggsound}. 
For example, when we hear a cat meow, we pay more attention to the cat than other objects due to the strong association between the meow and the cat.
Inspired by this potential interaction of auditory and visual signals, the cross-modal studies of vision and hearing have attracted numerous researchers, such as the audio-visual correspondence~\cite{2016soundnet,2017looklisten}, which only aims to match visual images and audio signals to the same scene, and sound source localization~\cite{2020_discriminative,2020_MSSL,2021localizingSSL} which further seeks to locate the vocal visible regions. 
However, they have only focused on audio-visual tasks at the image or region levels, lacking pixel-level annotations.
Recently, AVSBench~\cite{2022AVS, 2023AVSS} integrates audio signals into video segmentation, called Audio-Visual Segmentation~(AVS), which comprises two benchmarks:
1)~\emph{AVSBench-object}, which includes single source sound segmentation~(S4) and multiple sound sources segmentation~(MS3); 
2) \emph{AVSBench-semantic}, which further extends audio-visual semantic segmentation~(AVSS) based on AVSBench-object. 
 
Given that audio-visual segmentation is a burgeoning field that spans both audio and visual modalities, it presents a non-trivial task. 
Generally, when performing AVS, a cross-modal segmentation task involving video, there are mainly three challenges: (1) AVS contains audio and visual modalities, thus demanding explicit alignment of sequential audio features to spatial pixel-level activations; (2) AVS involves temporal information where the state of the current frame is dynamically affected by historical frames, therefore, exploring the correlation between adjacent frames is essential; (3) AVS includes an image segmentation task; compared to 1D audio signals, 2D image signals have more redundant information, which is prone to be affected by background noise, thus requiring precise extraction of features from the image. To resolve the issue (1), the prevailing methods~\cite{2023_ICCV_multimodal,2023AV-SAM,2023avsegformer,2023_bavs, ling2023hear} employ matrix multiplication and modified cross-attention module to encode pixel-wise audio-visual interaction.
Though impressive, these designs ignore the temporal dependence of adjacent frames that have been proven to be important for AVS.
To solve this problem, some methods~\cite{2022AVS,2023improving,2023catr} partition temporal relations into consideration to explore both issues (1) and (2) simultaneously since AVS is a cross-modal video task.
Nevertheless, their approaches rely too much on implicit inter-frame relations, leading to inaccurate associations.
Regarding issue~(3), CSMF~\cite{2023_leveraging} leverages frozen large foundation models to extract pure visual features for AVS. However, it independently tackles the audio and visual signals by naively combining several existing foundation models, resulting in sub-optimal performance.

To this end, we present \textbf{COMBO}, a novel audio-visual transformer framework for AVS. According to the three issues mentioned above, COMBO simultaneously considers modality, temporal, and pixel levels by introducing their bilateral entanglements, as shown in Fig.~\ref{fig:figure_task}. 
Specifically, nature itself has many bilateral relations. For example, in electricity and magnetism, due to their intrinsic correlation, the change of current causes the change of magnetic field, and vice versa. Motivated by this, we refer to this bilateral relationship of mutual influence as entanglement. 

In this work, we explore three potential bilateral entanglements: pixel entanglement, modality entanglement, and temporal entanglement. 
Pixel entanglement refers to the interdependent relationship between an image and its corresponding mask. 
Since background noise in the image leads to inaccuracies in the image-to-mask prediction process, it is essential to utilize external masks from the foundation models to entangle the input image to assist the model.
Therefore, we construct a Siam-Encoder Module~(SEM) as a visual feature extractor to facilitate more precise visual features, which can liberate from the constraints of foundation models than other methods~\cite{2023AV-SAM,2023gavs,2023_bavs}.
Besides, as for the alignments of the audio and visual signals, we explore the modality entanglement between audio and visual components to amplify the efficiency of cross-modal matching.
Contrary to existing single-fusion methods \cite{2022AVS, 2023avsegformer}, we believe that the cooperation between the two modalities can produce a positive effect. 
Inspired by \cite{2022_CVPR_GLIP}, we initially propose a potent and memory-efficient bidirectional audio-visual fusion module called Bilateral-Fusion Module~(BFM). 
Our BFM amplifies the spatial awareness of visual features relevant to sounding objects and strengthens the attention of audio signals embodying visual targets.
Moreover, the audio-visual tasks contain a solid temporal entanglement.
Thus, we design an adaptive inter-frame consistency loss to better harness this inherent characteristic.

Our main contributions can be summarized as follows:
\begin{itemize}

\item We propose a Siam-Encoder Module (SEM) that transfers the knowledge of the foundation model for mining the potential pixel entanglement. 

\item We propose a Bilateral-Fusion Module~(BFM) to take full advantage of the potential of both audio and visual modalities by exploring the modality entanglement.

\item We propose an adaptive inter-frame consistency loss based on the inherent coherence of audio-visual tasks for enhanced temporal entanglement.

\item We show that COMBO significantly outperforms existing state-of-the-art approaches in the challenging AVSBench-object and AVSBench-semantic datasets.  
\end{itemize}

\begin{figure*}[t!]
    \centering
    \includegraphics[width=2.0\columnwidth]{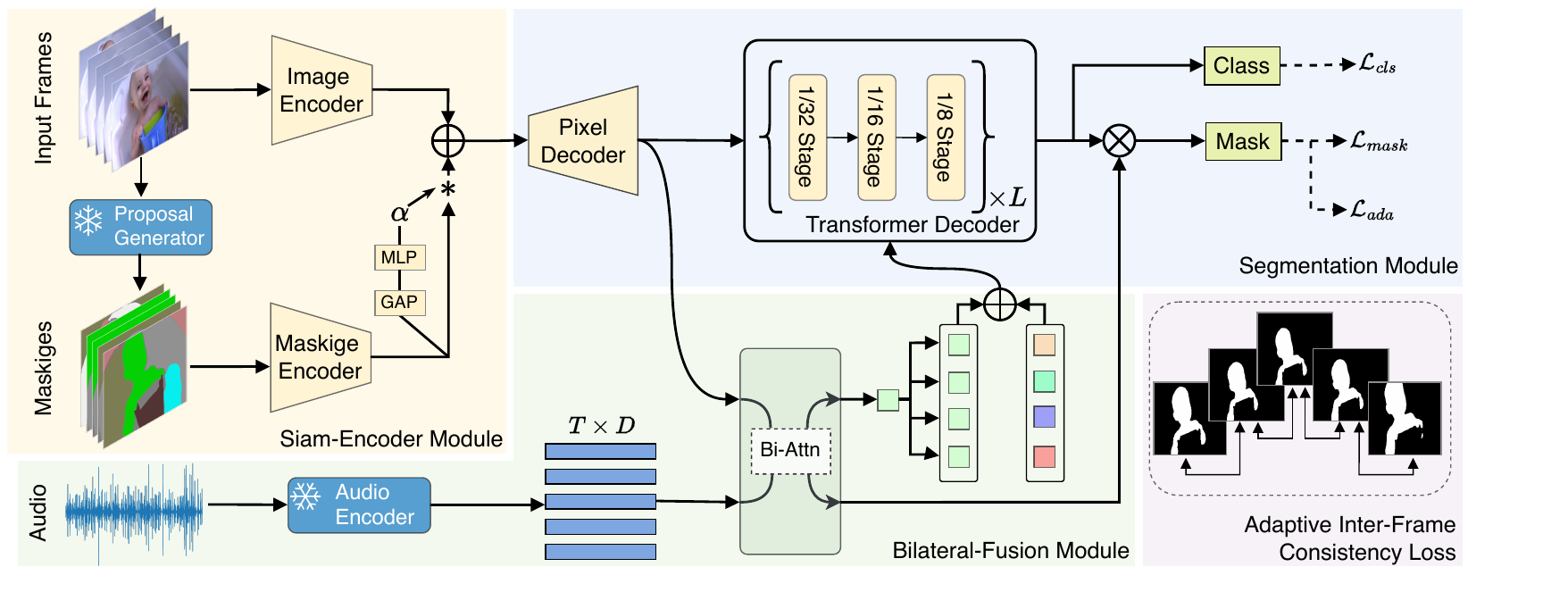}
    \vspace{-0.2cm}
    \caption{
    Overview of the proposed COMBO. COMBO adopts a novel audio-visual transformer framework specifically for audio-visual segmentation.
    Aiming at multi-oder bilateral entanglement, our method is composed of three independent modules. 
    (1) We introduce the Siam-Encoder Module, which is designed for the exploration of pixel entanglement.
    (2) To integrate the entanglement of audio and visual signals, we propose a Bilateral-Fusion Module.
    (3) Given the inherent characteristics of temporal entanglement, we construct an adaptive inter-frame consistency loss in the segmentation module to enhance the consistency of the output.
    } 
    \label{fig:figure_frame}
    \vspace{-0.5cm}
\end{figure*}

\section{Related Work}

\subsection{Sound Source Localization}
Sound source localization aims to estimate the position of a sound source in a video sequence, which is the most related task to the audio-visual segmentation task. 
LVS~\cite{2021localizingSSL} utilizes a hard-mining strategy and a contrastive learning mechanism to discriminate challenging image fragments.
DSOL~\cite{2020_discriminative} executes class-aware sounding object localization from mixed sound, which initially focuses on learning robust object representations from single-source localization.
MSSL~\cite{2020_MSSL} localizes multiple sound sources in unconstrained videos without pairwise sound-object annotations. 
This approach involves the development of a two-stage learning framework, followed by the execution of cross-modal feature alignment.
The pioneering methods in these areas have significantly inspired our research on AVS.

\subsection{Semantic Segmentation}
Semantic segmentation is a fundamental task that requires pixel-level classification.
Early researchers take the Fully Convolutional Networks~(FCN)~\cite{2015_FCN} as the dominant approach and focus on aggregating long-range context in the feature map.
PSPNet~\cite{2017_PSPNet} performs spatial pyramid pooling at several grid scales.
DeepLab~\cite{2018_deeplab,2017_rethinking} utilizes atrous convolutions with different atrous rates.
Furthermore, some methods~\cite{2021_segformer, 2021_Segmenter, 2021_SETR} replace traditional convolutional backbone with transformer-based architectures.
MaskFormer~\cite{cheng2021per} and Mask2Former~\cite{cheng2022masked} propose a mask classifier with learnable queries and specialized designs for mask prediction.
OneFormer~\cite{jain2023oneformer} presents a universal image segmentation framework that unifies segmentation with a multi-task train-once design.
Recently, a series of SAM models~\cite{2023_SAM_ICCV,2023_mobile_sam,2023_semanticsam,2023_lisa} propose to build a foundation model for promptable segmentation with strong generalization.
Given that the AVS task entails segmentation, these studies have significantly contributed to our work.

\subsection{Audio-Visual Segmentation}
AVS is an emerging task that aims to locate sounding sources by predicting pixel-wise maps and attracts many researchers~\cite{2023AVSS,2023avsegformer,2023_bavs,2023catr,2023gavs,2023_ICCV_multimodal,2023AV-SAM, 2023diffusionAVS}. 
AVSBench~\cite{2023AVSS} first constructs the audio-visual segmentation benchmark and proposes a temporal pixel-wise audio-visual interaction module (TPAVI) to inject audio semantics as guidance for the visual segmentation process. 
AVSegformer~\cite{2023avsegformer} proposes a transformer architecture that introduces audio features into the transformer decoder, enabling the network to attend to interested visual features selectively. 
CATR~\cite{2023catr} proposes a combinatorial dependence fusion approach that comprehensively accounts for the spatial-temporal dependencies of audio-visual combination.
Some methods~\cite{2023improving,2023diffusionAVS,2023_ICCV_multimodal} take advantage of the generative manners with latent diffusion model or variational auto-encoder to address AVS task.
In addition, AV-SAM~\cite{2023AV-SAM}, GAVS~\cite{2023gavs}, and BAVS~\cite{2023_bavs} utilize the large foundation model to bootstrap audio-visual segmentation.
Different from the above methods~\cite{2022AVS, 2023avsegformer,2023gavs,2023_bavs}, our proposed COMBO rethinks AVS from bilateral relations of three entanglements, which enhances the model's representation ability by exploring the pixel, modality and temporal inherent relationships.

\section{Method}

\subsection{Bilateral Visual Features Extraction}  \label{subsec:siam-network}
As illustrated in Fig.~\ref{fig:figure_frame}, our method initiates with the extraction of visual features primarily because the audio-visual segmentation (AVS), as a dense prediction task, exhibits extensive pixel entanglement in visual perception.
Recent studies~\cite{2023_semanticsam,2023_mobile_sam,2023_lisa} have demonstrated that the Segment Anything Model~\cite{2023_SAM_ICCV} exhibits robust generalization performance in segmentation tasks. 
Consequently, transferring the impressive capabilities of the foundation model to more complex visual tasks, such as AVS, presents an intriguing and valuable research question.
The extension, however, is not straightforward.
Although some methods~\cite{2023_leveraging,2023AV-SAM} attempt to fine-tune or concatenate the pre-trained SAM model for AVS, the limited capacity of the frozen foundation model restricts its performance to address the AVS task.
Additionally, the AVS task aims to predict all sound targets per pixel, whereas the SAM model is only capable of generating class-agnostic masks without any audio guidance, thus demonstrating a significant disparity. 
Therefore, transferring the foundation model's knowledge to the AVS task presents a tough challenge.


\noindent \textbf{Maskige as Prior Knowledge.}
To solve the aforementioned issues, we believe that a feasible strategy is to incorporate the knowledge of the foundation model into visual features as a pixel entanglement, which is memory-efficient.
Specifically, as depicted in the left area of Fig.~\ref{fig:figure_frame}, we introduce a Maskige generator. 
Given the input image $x \in \mathbb{R}^{3 \times H \times W}$ and the frozen foundation model, one can derive the class-agnostic masks $c \in \mathbb{R}^{K \times H \times W}$, where $K$ represents the number of potential targets.
It is essential to highlight that $K$ is dynamic and varies with input images.
Thus, we first amplify the quantity of class-agnostic masks from $K$ to $N$ with zero masks and obtain a series of binary masks, where $N$ is predetermined.
However, given that the output of SAM is a series of binary masks, it is difficult to integrate them into visual features. 
Consequently, inspired by \cite{2023_CVPR_generative}, we introduce Maskige $m \in \mathbb{R}^{3 \times H \times W}$, which shares the exact dimensions as the input image, to integrate prior knowledge better.
To efficiently generate Maskige $m$, we employ a random color encoding function $\mathcal{X}(\cdot):\mathbb{R}^{N} \rightarrow \mathbb{R}^{3}$ that is capable of transforming the binary masks $c \in \mathbb{R}^{N \times H \times W}$ into Maskige $m \in \mathbb{R}^{3 \times H \times W}$ without extra training.
Specifically,  $\mathcal{X}$ is designed to enhance the distinguishability of the Maskige and can be regarded as a linear layer $\mathcal{X}(c) = c A$, where $A \in \mathbb{R}^{N \times 3}$.
To facilitate offline inference using the Maskige generator, the value of $A$ is manually set appropriately without additional training. More details are in Appendix.

\begin{figure}[t!]
    \centering
    \includegraphics[width=1.0\columnwidth]{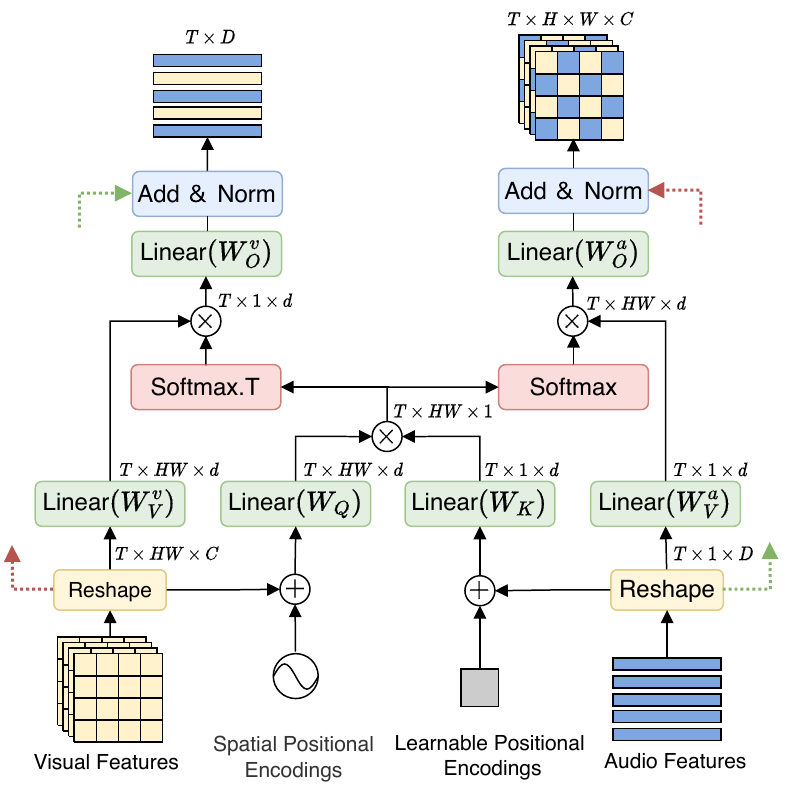}
    \vspace{-0.6cm}
    \caption{
    Illustration of Bilateral-Fusion Module~(BFM).
    We input both visual and image signals, which are subsequently processed through bilateral attention to yield the fused visual and image features respectively. We omit the subscripts of $H$ and $W$ for better understanding. 
    For enhanced visibility, the dashed line indicates a skip connection.
    Best viewed in color.} 
    \label{fig:figure_fusion}
    \vspace{-0.5cm}
\end{figure}
\noindent \textbf{Siam-Encoder Module (SEM)}.
To incorporate the image-like Maskiges as prior knowledge into input frames, we propose intertwining the features of Maskige and visual elements during the feature extraction stage.
Accordingly, we design a Siam-Encoder Module, as depicted in Fig.~\ref{fig:figure_frame}.
This module encompasses an Image encoder $E_v$ and a Maskige encoder $E_m$, sharing a common framework.
More precisely, for a short video clip with $T$ frames $I \in \mathbb{R}^{T \times 3 \times H \times W}$, the Maskiges can be generated using the Maskige generator, resulting in $M \in \mathbb{R}^{T \times 3 \times H \times W}$.
Subsequently, we extract multiple output features from both the image encoder and Maskige encoder, respectively. This process can be defined as follows:
\begin{equation}
  F_{visual} = E_v (I;\theta_v), \; F_{visual} \in \{F_{v_i} \}_{i=1}^4, 
  \label{eq:visual_encoder}
\end{equation}
\begin{equation}
  F_{maskige} = E_m (M;\theta_m), \; F_{maskige} \in \{F_{m_i}\}_{i=1}^{4},
  \label{eq:maskige_encoder}
\end{equation}
in which  $F_{v_i}$  and  $ F_{m_i}  \in \mathbb{R}^{T \times \frac{H}{2^{i+1}} \times \frac{W}{2^{i+1}} \times C_i}$ . $C_i$ represents the dimension of the $i$-th stage output features. 
To integrate the Maskige features into COMBO, we introduce channel-weighted blocks that augment the original visual features. The formula can be written as follows:
\begin{equation}
  F_{v_i} = F_{m_i}(\text{GAP}(F_{m_i}) W )   + F_{v_i}, \; i = \{1,2,3,4\},
  \label{eq:weight_add}
\end{equation}
where GAP stands for global average pooling, and $W \in \mathbb{R}^{C_i \times C_i}$ represents the linear weight. 
For simplicity, the bias is omitted in this context.
After obtaining the Maskiges as prior information to boost pixel-level entanglement with visual features, the next critical aspect is exploring the modality entanglement between audio and visual signals.

\subsection{Audio-Visual Bilateral Fusion}  \label{subsec:siam-fusion}
The relationship between any two modalities can be characterized as bilateral entanglement.
For instance, an image can be described in text, and sound is inextricably linked with its visual counterpart.
This entanglement among these distinct modalities provides an invaluable resource for researchers tackling multimodal tasks. 
Prior studies~\cite{2022AVS,2021_segformer} have overemphasized the influence of audio on visual features, thereby underestimating the significance of visual information to audio features.
To address this imbalance, we propose a Bilateral-Fusion Module~(BFM) in COMBO that surpasses a mere single fusion effect.

\noindent \textbf{Audio Feature Extraction.} 
For an audio clip corresponding to the input frames, we adopt VGGish \cite{2017_vggish} to extract audio features following~\cite{2022AVS}.
Firstly, the audio clip is resampled to yield a 16kHz mono output $A_{mono} \in \mathbb{R}^{N_{samples} \times 96 \times 64}$, where $N_{samples}$ is related to the duration of the audio.
Then, a short-time Fourier transform is performed to yield a mel spectrum, denoted as $A_{mel} \in \mathbb{R}^{T \times 96 \times 64}$.  
Finally, the mel spectrum is subsequently fed into the VGGish model, resulting in the extraction of audio features $F_{a} \in \mathbb{R}^{T \times D}$, where $T$ denotes the number of frames and  $D$ represents dimension of the audio.
\begin{figure}[t!]
    \centering
    \includegraphics[width=1.0\columnwidth]{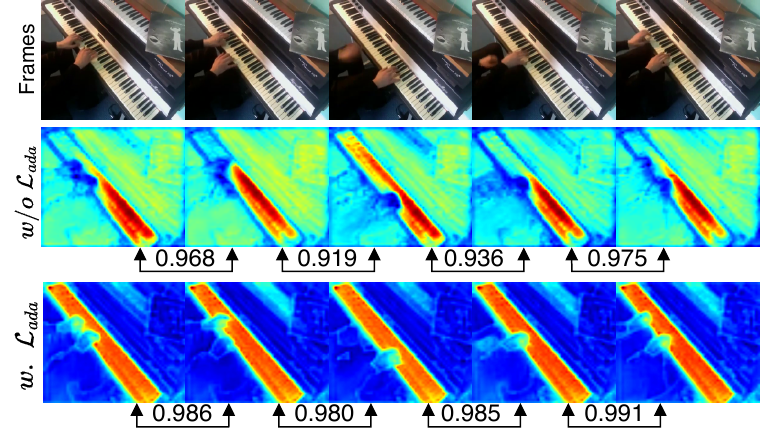}
    \vspace{-0.5cm}
    \caption{
    Illustration of the impact on Adaptive Inter-frame Consistency Loss.
    We visualize the heat map of the predicted masks without and with the consideration of $\mathcal{L}_{ada}$ based on the S4 subset.
    The results indicate that implementing $\mathcal{L}_{ada}$ promotes superior interframe consistency.
    Best viewed in color.} 
    \label{fig:figure_cosine}
    \vspace{-0.5cm}
\end{figure}

\noindent \textbf{Bilateral-Fusion Module~(BFM).}
We initially employ the pixel decoder~\cite{2021_DN_DETR} to gradually upsample visual features $F_{v_{i}}$ derived from the SEM to further generate high-resolution per-pixel embeddings $P_{i} \in \mathbb{R}^{T \times \frac{H}{2^{i+1}} \times \frac{W}{2^{i+1}} \times C }, \; i \in \{1,2,3,4\}$, where $C$ denotes the output channel.
Then, we design a Bilateral-Fusion Module~(BFM) for constructing a bidirectional audio-visual mapping to assist with segmenting the sounding objects.

As shown in Fig.~\ref{fig:figure_fusion}, our BFM utilizes audio features $F_{a} \in \mathbb{R}^{T \times D}$,  in conjunction with the largest pixel-level embeddings $P_{1} \in \mathbb{R}^{T \times H_1 \times W_1 \times C }$ as inputs which can propagate ample fine-grained semantic information to audio features. Here, $H_1 = H/4,; W_1 = W/4$.
In order to incorporate both two signals, bilateral attention is designed within our BFM.
Specifically, we initially add fixed sine spatial positional encodings and learnable positional encodings to $P_1$ and $F_a$, respectively.
Next, in order to integrate audio-visual modalities in a memory-efficient way, our BFM comprises four point-wise linear layers that map $P_1$ and $F_a$ to intermediate representations with dimension $d$. 
These representations share queries and keys with queries $Q = P_1 W_Q $, keys $K = F_a W_K $, visual values $ V_{v} = P_1 W_{V}^{v}$, and audio values $V_a = P_1 W_V^a$.
Following the mapping process, the bilateral attention is as follows:
\begin{equation}
P_1 = \text{Softmax}(QK^T / \sqrt{d})V_a + P_1,
\end{equation}
\begin{equation}
F_a = \text{Softmax}((QK^T / \sqrt{d})^T)V_v + F_a,
\end{equation}
where $d$ is the embedding dimension. And $(QK^T / \sqrt{d})$ is calculated only once, which is more efficient.

After BFM, we proceed by expanding fused audio features $F_a$ added with learnable embeddings within the transformer decoder as object queries.
Additionally, following \cite{2022_Mask2Former}, we generate the output classes $O^{cls}$ by incorporating the per-pixel embeddings $P_4, P_3, P_2$ into the transformer decoder.
We acquire the predicted masks $O^{mask}$ by multiplying the output embeddings from the transformer decoder with the fused embedding $P_1$.

\subsection{Mining Temporal Relationships}  \label{subsec:loss}

\noindent \textbf{Adaptive Inter-frame Consistency Loss.}  
Temporal always implies a bilateral relationship in nature.
For instance, in video clips, one can deduce the scenario of the current frame based on the past frame. Similarly, it is also feasible to predict the future frame based on the current frame. 
This interactive relations among frames can be construed as a type of temporal entanglement.
To take advantage of this potential temporal entanglement, we introduce an adaptive inter-frame consistency loss for AVS.
The similarity score for each successive frame can be calculated as follows:
\begin{equation}
\mathcal{S}_{t : t+1} = \text{cos}(O_t^{mask}, O_{t+1}^{mask}),
\end{equation}
where $O_{t}^{mask}$ refers to the predicted masks at frame $t$, and $\text{cos}(\cdot)$ symbolizes the cosine similarity function. 
The term $\mathcal{S}_{t:t+1}$ represents the similarity score between frames $t$ and $t+1$.
As illustrated in Fig.~\ref{fig:figure_cosine}, it is evident that a significant similarity exists between distinct frames.
Therefore, to leverage this prior information, we propose an adaptive inter-frame consistency loss, formulated as follows:
\begin{equation}
\mathcal{L}_{ada} =\sum_{t=1}^{T-1} \exp(\mathcal{S}_{t:t+1}-1)(1-\mathcal{S}_{t:t+1}),
\end{equation}
where $\exp(\mathcal{S}_{t:t+1}-1)$ represents adaptive weight. 
When the disparity between adjacent frames is substantial, the adaptive weight item is minimal, aligning with intuition.

\subsection{Training and Inference}  \label{subsec:total_loss}

\noindent \textbf{Overall Training Loss.}
The comprehensive training loss comprises three components: classification loss, mask loss, and adaptive inter-frame consistency loss, as previously discussed.
The classification loss is formulated by a cross-entropy loss, denoted as $\mathcal{L}_{cls}= \mathcal{L}_{ce}$.
The mask loss integrates the binary cross-entropy loss and the dice loss~\cite{2016_dice_loss}, and is depicted as $\mathcal{L}_{mask} = \mathcal{L}_{bce} + \mathcal{L}_{dice}$.
Considering that in the AVS task, the ratio of segmented objects to the total image area is relatively small, employing dice loss allows the model to better concentrate on the foreground and minimizes distraction from the background.
The overall training loss is expressed as follows:
\begin{equation}
\label{eq:total_loss}
\mathcal{L} = \lambda_{cls} \mathcal{L}_{cls} + \lambda_{mask} \mathcal{L}_{mask} 
 + \lambda_{ada} \mathcal{L}_{ada}, 
\end{equation}
where $\lambda_{cls}$, $\lambda_{mask}$, and $\lambda_{ada}$ are hyperparameters.
More details about the $\lambda$ parameters are available in Sec.~\ref{sec:settings}.

\noindent \textbf{Semantic Inference.}  
After obtaining the predicted embeddings $O^{cls} \in \mathbb{R}^{T \times N_q \times (K_c+1)}$ and binary masks $O^{mask} \in \mathbb{R}^{T \times N_q \times H \times W}$, where $K_c$ represents the total number of object classes and $N_q$ is the number of object queries, we employ the identical post-processing as in \cite{2022_Mask2Former} to yield the final semantic segmentation outputs.
Specifically, we first calculate the output mask with classes $O = O^{cls} \times O^{mask} \in \mathbb{R}^{T \times (K_c+1) \times H \times W }$.
Then, we execute $\arg \max$ and discard the \emph{no object} class to obtain the ultimate results.

\begin{table}[t!]
    \small
    \centering
    \vspace{-0.2cm}
    \setlength{\tabcolsep}{4pt} 
    \renewcommand{\arraystretch}{1.0} 
    \begin{tabular}{llcccc} 
        \toprule 
        \multirow{2}*{\textbf{Method}}  & \multirow{2}*{\textbf{Backbone}} & \multicolumn{2}{c}{\textcolor{blue}{\textbf{S4}}} & \multicolumn{2}{c}{\textcolor{red}{\textbf{MS3}}}  \\
        \cmidrule(lr){3-4} \cmidrule(lr){5-6}
        & & $\mathcal{M_J}$ & $\mathcal{M_F}$ & $\mathcal{M_J}$ & $\mathcal{M_F}$ \\
        \midrule 
        LVS~\cite{2021localizingSSL}  & ResNet-18 & 37.9 & 51.0 & 29.5 & 33.0   \\
         MSSL~\cite{2020_MSSL}  & ResNet-18 & 44.9 & 66.3 & 26.1 & 36.3   \\
         3DC~\cite{2020_3dc}  & ResNet-152 & 57.1 & 75.9 & 36.9 & 50.3   \\
         SST~\cite{2021_SST}  & ResNet-101 & 66.3 & 80.1 & 42.6 & 57.2   \\
         iGAN~\cite{2022_iGAN}  & ResNet-50 & 61.6 & 77.8 & 42.9 & 54.4   \\
         LGVT\cite{2021_LGVT}  & Swin-B & 74.9 & 87.3 & 40.7 & 59.3   \\
        \midrule 
          {AVSBench~\cite{2023AVSS}}   & ResNet-50 & 72.8 & 84.8 & 47.9 & 57.8   \\
          & PVT-v2 & 78.7 & 87.9 & 54.0 & 64.5  \\
          {CSMF~\cite{2023_leveraging}}   & ViT-B & 58.0 & 67.0 & 34.0 & 44.0   \\ 
         {AVS-BiGen~\cite{2023improving}}  & ResNet-50 & 74.1 & 85.4 & 45.0 & 56.8 \\
         & PVT-v2 & 81.7 & 90.4 & 55.1 & 66.8 \\
        {CATR~\cite{2023catr}}  & ResNet-50 & 74.8 & 86.6 & 52.8 & 65.3  \\
          & PVT-v2 & 81.4 & 89.6 & 59.0 & 70.0 \\
         {DiffusionAVS~\cite{2023diffusionAVS}}  & ResNet-50 & 75.8 & 86.9 & 49.8 & 58.2  \\
          & PVT-v2 & 81.4 & 90.2 & 58.2 & 70.9 \\
         {ECMVAE~\cite{2023_ICCV_multimodal}}  & ResNet-50 & 76.3 & 86.5 & 48.7 & 60.7  \\
          & PVT-v2 & 81.7 & 90.1 & 57.8 & 70.8 \\
         {BAVS~\cite{2023_bavs}}  & ResNet-50 & 78.0 & 85.3 & 50.2 & 62.4  \\
         & PVT-v2 & 82.0 & 88.6 & 58.6 & 65.5  \\
         {AVSegFormer~\cite{2023avsegformer}}  & ResNet-50 & 76.5 & 85.9 & 49.5 & 62.8  \\
          & PVT-v2 & 82.1 & 89.9 & 58.4 & 69.3 \\
          \midrule 
        \vspace*{-4pt}
         \textbf{COMBO~(ours)}  & \textbf{ResNet-50} & \textbf{81.7} & \textbf{90.1} & \textbf{54.5} & \textbf{66.6}  \\
         \scriptsize & \scriptsize & \scriptsize \textbf{\textcolor{expgreen}{ (+3.7)}} & \scriptsize \textbf{\textcolor{expgreen}{ (+4.8)}} & \scriptsize \textbf{\textcolor{expgreen}{ (+2.7)}} & \scriptsize \textbf{\textcolor{expgreen}{ (+1.3)}}  \\
        \vspace*{-4pt}
         & \textbf{PVT-v2} & \textbf{84.7} & \textbf{91.9} & \textbf{59.2} & \textbf{71.2}  \\
         \scriptsize & \scriptsize & \scriptsize \textbf{\textcolor{expgreen}{ (+2.6)}} & \scriptsize \textbf{\textcolor{expgreen}{ (+2.0)}} & \scriptsize \textbf{\textcolor{expgreen}{ (+0.2)}} & \scriptsize \textbf{\textcolor{expgreen}{ (+1.2)}}  \\
        \bottomrule 
    \end{tabular}
    \vspace{-0.2cm}
    \caption{Quantitative comparison results of different methods on AVSBench-object (Single-source, \textcolor{blue}{S4}; Multi-source, \textcolor{red}{MS3}). We use the same backbones (ResNet-50 and PVT-v2) to demonstrate that our method outperforms other methods significantly.}
    \label{tab:exp_main}
     \vspace{-0.2cm}
\end{table}

\section{Experiments}

\subsection{AVSBench Datasets} \label{sec:dataset}
We evaluate our proposed method on the AVSBench dataset~\cite{2023AVSS}, which consists of two scenarios: AVSBench-object and AVSBench-semantic.

\begin{table}[t!]
    \centering
    \small
    \vspace{-0.2cm}
    \setlength{\tabcolsep}{11pt} 
    \renewcommand{\arraystretch}{1.0} 
    \begin{tabular}{llccc} 
        \toprule 
        \multirow{2}*{\textbf{Method}}  & \multirow{2}*{\textbf{Backbone}} & \multicolumn{2}{c}{\textbf{AVSS}} \\
        \cmidrule(lr){3-4} 
        & & $\mathcal{M_J}$ & $\mathcal{M_F}$  \\
        \midrule 
        3DC~\cite{2020_3dc}  & ResNet-18 & 17.3 & 21.6  \\
        AOT~\cite{2021_AOT} & ResNet-50 & 25.4 & 31.0   \\
        \midrule  
        AVSBench~\cite{2023AVSS}  & ResNet-50 & 20.2 & 25.2    \\
         & PVT-v2 & 29.8 & 35.2  \\
        BAVS~\cite{2023_bavs}  & ResNet-50 & 24.7 & 29.6  \\
         & PVT-v2 & 32.6 & 36.4   \\
        AVSegFormer~\cite{2023avsegformer}  & ResNet-50 & 24.9 & 29.3  \\
         & PVT-v2 & 36.7 & 42.0  \\
         \midrule 
        \vspace*{-4pt}
         \textbf{COMBO~(ours)}  & \textbf{ResNet-50} & \textbf{33.3} & \textbf{37.3}   \\
         \scriptsize & \scriptsize & \scriptsize \textbf{\textcolor{expgreen}{ (+8.4)}} & \scriptsize \textbf{\textcolor{expgreen}{ (+8.0)}}   \\
        \vspace*{-4pt}
          & \textbf{PVT-v2} & \textbf{42.1} & \textbf{46.1}   \\
        \scriptsize & \scriptsize & \scriptsize \textbf{\textcolor{expgreen}{ (+5.4)}} & \scriptsize \textbf{\textcolor{expgreen}{ (+4.1)}}   \\
        
        \bottomrule 
    \end{tabular}
    \vspace{-0.2cm}
    \caption{Quantitative comparison results on AVSBench-semantic.}
    \label{tab:exp_avss}
    \vspace{-0.1cm}
\end{table}


\begin{table}[t!]
    \centering
    \small
    \vspace{-0.2cm}
    \setlength{\tabcolsep}{6pt} 
    \renewcommand{\arraystretch}{0.9} 
    \begin{tabular}{lcccc} 
        \toprule 
         \multirow{2}*{\textbf{Module}} & \multicolumn{2}{c}{\textbf{S4}}  & \multicolumn{2}{c}{\textbf{AVSS}} \\
        \cmidrule(lr){2-3}  \cmidrule(lr){4-5}  
        &  $\mathcal{M_J}$ & $\mathcal{M_F}$ & $\mathcal{M_J}$ & $\mathcal{M_F}$  \\
        \midrule
        \textbf{COMBO} & \textbf{81.7} & \textbf{90.1} & \textbf{33.3} & \textbf{37.3} \\ 
        $w/o$ Siam-Encoder & 80.6  & 88.7  & 31.9 & 35.7 \\
        $w/o$  Bilateral-Fusion &  81.1 & 89.9 & 33.1 & 36.7  \\
        $w/o$ Inter-Frame Loss&  81.0 & 89.8  & 33.0 & 37.1  \\
 
        \bottomrule 
    \end{tabular}
    \vspace{-0.2cm}
    \caption{Ablation study of the various modules included in COMBO. 
    We sequentially remove our proposed modules and compare their performance.}
     \vspace{-0.4cm}
    \label{tab:exp_ablation_all}
\end{table}

\noindent \textbf{AVSBench-object.}
AVSBench-object \cite{2022AVS} is an audio-visual dataset specifically designed for sound target segmentation. 
AVSBench-object includes two scenarios based on the number of audio sources in each frame: single sound source segmentation (S4) and multiple sound source segmentation (MS3).
The S4 scenario incorporates 4,932 videos, with the ratio of train/validation/test split ratio configured at 70/15/15.
This scenario is trained in a semi-supervised manner, wherein each video comprises five frames, but annotation during training is limited to the first frame only.
Conversely, the MS3 scenario is characterized by multiple sound sources, including 424 videos, and maintains the same split ratios as in the S4 scenario.
This scenario, unlike S4, employs a fully supervised training approach with all five frames being annotated.

\noindent \textbf{AVSBench-semantic.} 
AVSBench-semantic~\cite{2023AVSS} is an extension to AVSBench-object that incorporates additional semantic labels for the purpose of enhancing audio-visual semantic segmentation~(AVSS).
AVSBench-semantic includes a set of new multi-source videos as well as the original AVSBench-object videos, cumulatively accounting for a total of 11,356 videos spanning 70 categories. These videos are allocated as follows: 8,498 for training, 1,304 for validation, and 1,554 for testing.
Additionally, the original videos have only been enhanced with semantic information, maintaining the same frames as prior. 
However, the new videos have been extended to $10$ frames, increasing the difficulty due to the inclusion of extended audio-visual sequences.

\subsection{Experimental Setup} \label{subsec:imple_details}
\noindent \textbf{Implementation Details.} \label{sec:settings}
For fair comparison, we adopt the ImageNet pre-trained ResNet-50~\cite{2016_Resnet} and Pyramid Vision Transformer (PVT-v2)~\cite{2022_pvt} as the visual siam-encoders.
All input frames are resized to $224 \times 224$.
In terms of the proposal generator, we leverage the Semantic-SAM~\cite{2023_semanticsam} to obtain class-agnostic masks.
As for audio input, we take the Vggish encoder~\cite{2017_vggish} pre-trained on AudioSet~\cite{2023_Audioset} to extract audio features.
Following \cite{2022_Mask2Former}, the Multi-Scale Deformable Attention Transformer (MSDeformAttn) is our default pixel decoder.
Besides, we adopt the standard transformer decoder, with $L=3$ (i.e., a total of 9 layers) and $N_q=100$ as the default.
The hyperparameters are set as $\lambda_{cls}=2$ and $\lambda_{mask}=5$. 
For the inter-frame consistency loss, we set $\lambda_{ada}=10$ for the AVSBench-object dataset, $\lambda_{ada}=5$ for the AVSBench-semantic dataset due to longer frames per video. 
We calculate the inter-frame consistency loss only in the intermediate transformer decoder layer to prevent an over-dependence on temporal information.
All models are trained using the Adam optimizer with a learning rate of $1e-4$ and weight decay of $0.05$.
We train the S4 and AVSS subsets for 90k and MS3 for 20k iterations with a batch size of 8 on a single A100 40GB GPU.

\noindent \textbf{Metrics.}
Folloing~\cite{2022AVS}, we adopt two metrics to verify the effectiveness, namely, Jaccard index $\mathcal{J}$ and F-score $\mathcal{F}$. 
 $\mathcal{J}$ computes the intersection over union (IoU) of the predicted segmentation and the ground truth mask.
$\mathcal{F}$ considers both precision and recall, which is represented as 
$\mathcal{F}_\beta = \frac{(1+\beta^2) \times \text{presion} \times \text{recall} }{\beta^2 \times \text{precision} + \text{recall}}$, where $\beta^2$ is set at $0.3$.
In our experiment, we use $\mathcal{M_J}$ and $\mathcal{M_F}$ to denote the mean metrics across the entire dataset.

\begin{table}[t!]
    \centering
    \small
    \vspace{-0.2cm}
    \setlength{\tabcolsep}{5pt} 
    \renewcommand{\arraystretch}{0.9} 
    \begin{tabular}{lccccc} 
        \toprule 
         \multirow{2}*{\textbf{SEM Module}}  & \multicolumn{2}{c}{\textbf{S4}}  & \multicolumn{2}{c}{\textbf{AVSS}} & \multirow{2}*{\textbf{\#Params}}   \\
        \cmidrule(lr){2-3}  \cmidrule(lr){4-5}  
         & $\mathcal{M_J}$ & $\mathcal{M_F}$ & $\mathcal{M_J}$ & $\mathcal{M_F}$  \\
        \midrule
         $w.$ shared weights &81.5 & 90.2 & 31.1 & 34.4 & 135.5 \\ 
         $w/o$ shared weights &  \textbf{81.7} & \textbf{90.1} & \textbf{33.3} & \textbf{37.3}  &  158.9  \\
        \bottomrule 
    \end{tabular}
    \vspace{-0.2cm}
    \caption{Ablation study of Siam-Encoder Module~(SEM).  
    We explore two configurations: using shared weights and using separate weights. \#Params denotes the model parameters (M).}

    \label{tab:exp_ablation_share}
\end{table}

\begin{table}[t!]
    \centering
    \small
    \vspace{-0.2cm}
    \setlength{\tabcolsep}{10pt} 
    \renewcommand{\arraystretch}{0.9} 
    \begin{tabular}{lcccc} 
        \toprule 
         \multirow{2}*{\textbf{Fusion Mode}}  & \multicolumn{2}{c}{\textbf{S4}}  & \multicolumn{2}{c}{\textbf{AVSS}} \\
          \cmidrule(lr){2-3}  \cmidrule(lr){4-5}  
         & $\mathcal{M_J}$ & $\mathcal{M_F}$ & $\mathcal{M_J}$ & $\mathcal{M_F}$  \\
        \midrule
         $w/o$ fusion &  81.1 & 89.9 & 33.1 & 36.7 \\ 
         visual fusion only  & 81.3  &  89.6 & 32.6 & 36.4    \\
         audio fusion only   & 81.3  &  \textbf{90.2} & 33.2 & 36.7  \\
         \textbf{bilateral fusion}  & \textbf{81.7}  &  90.1  & \textbf{33.3} & \textbf{37.3}  \\
        \bottomrule 
    \end{tabular}
    \vspace{-0.2cm}
    \caption{Ablation study of fusion mode. 
    We conduct among four modes: no fusion, visual fusion only, audio fusion only, and our fusion mode.}
    \label{tab:exp_ablation_fusion}
    \vspace{-0.5cm}
\end{table}

\subsection{Main Results} 
We conduct experiments on AVSBench-object and AVSBench-semantic datasets.
As AVS is an emerging proposed problem recently introduced by \cite{2022AVS}, we compare our COMBO with some state-of-the-art methods from other related tasks, such as sound source localization (SSL)~\cite{2021localizingSSL,2020_MSSL}, video object segmentation (VOS)~\cite{2020_3dc,2021_SST,2021_AOT}, and salient object detection (SOD)~\cite{2022_iGAN,2021_LGVT}, all of which provide a comparative benchmark for our experiments.
As evidenced in Tab.~\ref{tab:exp_main}, COMBO demonstrates a substantial performance gap ($+9.8$ mIoU in S4; $+16.3$ mIoU in MS3) over other related methods, principally attributable to variances in setting specific task scenarios. 
We also compare our method against some recent state-of-the-art methods \cite{2022AVS,2023improving,2023catr,2023diffusionAVS,2023_ICCV_multimodal,2023_bavs,2023avsegformer} that have been explicitly designed for audio-visual segmentation settings.
On the AVSBench-object dataset, COMBO-R50 outperforms the current best performance by achieving 3.7 mIoU and 4.8 F-score improvements for S4 subset and 2.7 mIoU and 1.3 F-score improvements for MS3, while COMBO-PVT surpasses the top-performing method by 2.6 mIoU and 2.0 F-score for S4 and 0.2 mIoU and 1.2 F-score for MS3.

Besides, we compare the AVSBench-semantic dataset as displayed in Tab.~\ref{tab:exp_avss}, which presents a more challenging setting.
Both COMBO-R50 and COMBO-PVT achieve significant results, with 8.4 and 5.4 mIoU improvements and significant F-score enhancements of 8.0 and 4.1, respectively.
These experiments confirm that our COMBO model surpasses existing state-of-the-art methods across all sub-tasks, consequently setting a new benchmark for AVS.

\begin{table}[t!]
    \centering
    \small
    \vspace{-0.2cm}
    \setlength{\tabcolsep}{12pt} 
    \renewcommand{\arraystretch}{0.9} 
    \begin{tabular}{lcccc} 
        \toprule 
         \multirow{2}*{\textbf{Queries}}  & \multicolumn{2}{c}{\textbf{S4}}  & \multicolumn{2}{c}{\textbf{AVSS}} \\
        \cmidrule(lr){2-3}  \cmidrule(lr){4-5}  
         & $\mathcal{M_J}$ & $\mathcal{M_F}$ & $\mathcal{M_J}$ & $\mathcal{M_F}$  \\
        \midrule
         all  & 80.9 & 89.9 & 30.7 & 34.2     \\
         \textbf{add}  &  \textbf{81.7} & \textbf{90.1} & \textbf{33.3} & \textbf{37.3}  \\ 
        \bottomrule 
    \end{tabular}
    \vspace{-0.1cm}
    \caption{Ablation study of learnable queries.}
    \label{tab:exp_num_queries}
\end{table}
\begin{table}[t!]
    \centering
    \small
    \vspace{-0.2cm}
    \setlength{\tabcolsep}{13pt} 
    \renewcommand{\arraystretch}{0.9} 
    \begin{tabular}{ccccc} 
        \toprule 
         \multirow{2}*{$\lambda_{ada}$} & \multicolumn{2}{c}{\textbf{S4}}  & \multicolumn{2}{c}{\textbf{AVSS}} \\
        \cmidrule(lr){2-3}  \cmidrule(lr){4-5}  
        & $\mathcal{M_J}$ & $\mathcal{M_F}$ & $\mathcal{M_J}$ & $\mathcal{M_F}$  \\
        \midrule
         0 &  81.0 & 89.8 & 33.0 & 37.1 \\ 
         5  & 81.1 & \textbf{90.1} &  \textbf{33.3} & \textbf{37.3}   \\   
         10  & \textbf{81.7}  & \textbf{90.1} &  32.1 & 35.6   \\
         20  & 81.2  &  89.7 & 32.6 & 36.1  \\
        \bottomrule 
    \end{tabular}
    \vspace{-0.2cm}
    \caption{Ablation study of the adaptive inter-frame consistency loss. $\lambda_{ada}$ is the hyperparameter, while higher values constrain the output to be more similar.}
    \vspace{-0.3cm}
    \label{tab:exp_ablation_loss}
\end{table}

\subsection{Abaltion Study}
In this section, we conduct ablation studies to verify the effectiveness of each essential design in the proposed COMBO. 
Specifically, we adopt ResNet-50~\cite{2016_Resnet} as the backbone and carry out extensive experiments on the S4 and AVSS sub-tasks due to more videos in these tasks. Other training settings remain consistent with Sec.~\ref{subsec:imple_details}
\begin{figure*}[t!]
    \centering
    \includegraphics[width=2.05\columnwidth]{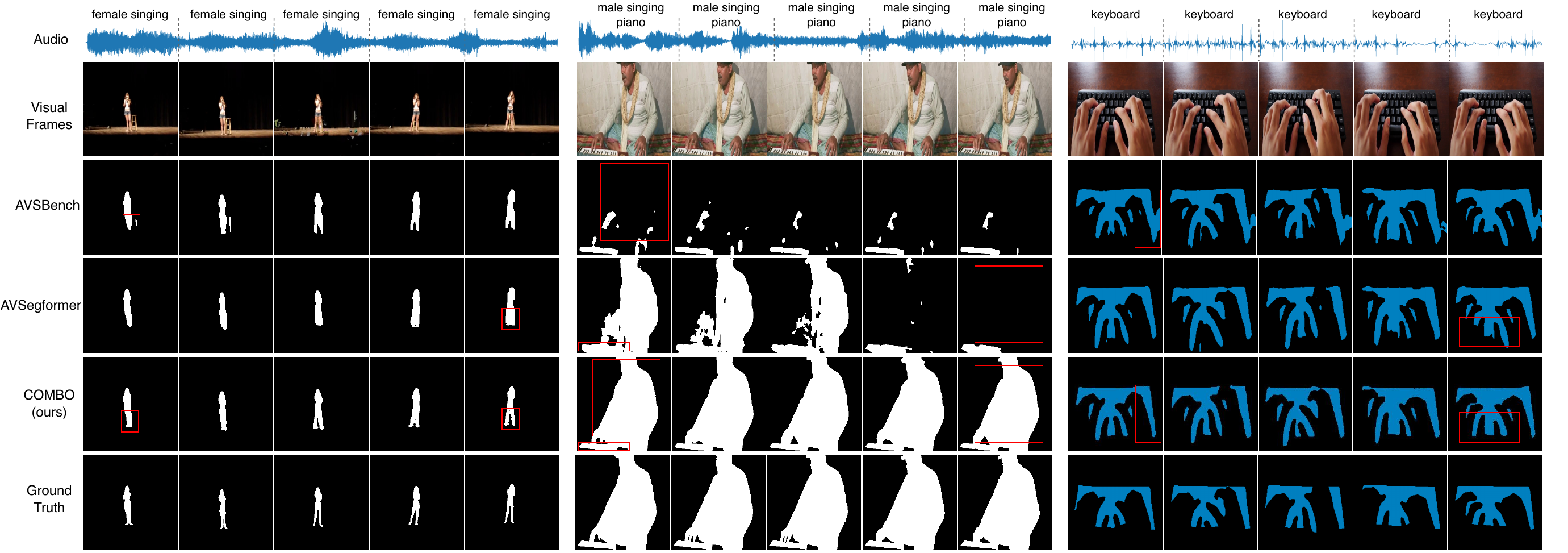}
    \vspace{-0.1cm}
    \caption{
    Comparison of Visual Examples on the AVSBench-object and AVSBench-semantic Datasets with AVSBench~\cite{2023AVSS} and AVSegformer~\cite{2023avsegformer}.
    Wherein the leftmost example is derived from the S4 subset, the middle example is from the MS3 subset, and the rightmost example is from the AVSS subset.
    Red bounding boxes highlight the specific regions for comparison. 
    } 
    \label{fig:figure_visualize}
    \vspace{-0.40cm}
\end{figure*}

\noindent \textbf{Component analysis of COMBO.} 
To validate the impact of our proposed method, we separately eliminate the Siam-Encoder Module~(SEM), Bilateral-Fusion Module (BFM), and adaptive inter-frame consistency loss.
As demonstrated in Tab.~\ref{tab:exp_ablation_all}, the results indicate that COMBO has demonstrated superior influence on SEM with 1.4 mIoU improvement, particularly on the AVSS subset.
Concurrently, our findings indicate that BFM is of substantial significance, demonstrating a performance enhancement of 0.6 mIoU over the S4 subset. More analysis are discussed later. 
In addition, we examine the effect of inter-frame consistency loss.
The results reveal that our loss function contributes to performance improvements with 0.7 mIoU on the S4 subset. More details are provided subsequently.

\noindent \textbf{Effects of Siam-Encoder Module~(SEM).} 
We first examine the significance of our SEM. 
Two designs are compared in Tab.~\ref{tab:exp_ablation_share}: one with share weights and another with separate weights.
When comparing shared and separate parameters, it is found that separable parameters can attain 0.2 mIoU on the S4 subset, but it is indeed more costly than the others.

\noindent \textbf{Effects of Bilateral-Fusion Module (BFM).} 
Moreover, we investigate the influence of our BFM.
We compared our modules using different variants, which include without any fusion, video-only fusion~(inject audio feature into visual), audio-only fusion~(inject visual feature to audio), and our bilateral fusion.
As demonstrated in Tab.~\ref{tab:exp_ablation_fusion}, bilateral fusion achieves a performance improvement of 0.6 mIoU compared to the model without any fusion.
It is worth noting that the transformer decoder inherently consists of a cross-attention function, potentially serving as a fusion process.
Our model also realizes a performance improvement of 0.4 mIoU compared to only audio or visual fusion.

\noindent \textbf{Effects of Audio Queries.} 
As demonstrated in Tab.~\ref{tab:exp_num_queries}, \emph{All} denotes the exclusive use of fused audio queries, and \emph{Add} signifies the combination of fused audio queries and learnable queries. We first expand the audio features to the exact dimensions as learnable queries, then compare the experiments with the exclusive use of fused audio queries and the combination of fused audio queries and learnable queries. 
The results show that queries with \emph{Add} have 0.8 and 2.6 mIoU improvements over \emph{All} on S4 and AVSS subsets.

\noindent \textbf{Effects of Adaptive Inter-frame Consistency Loss.} 
To validate the effect of $\mathcal{L}_{ada}$, we adopt different values of $\lambda_{ada}$.
As shown in Tab.~\ref{tab:exp_ablation_loss}, the experiments demonstrate that appropriate consistency constraints can enhance the model's performance.
The appropriate value $\lambda_{ada}$ can improve performance by 0.7 mIoU and 0.3 F-score for S4.
In addition, the heat map of the predicted masks has been visualized in Fig.~\ref{fig:figure_cosine}.
It is observable that the use of $\mathcal{L}_{ada}$ facilitates the generation of a more distinct boundary output.
Nevertheless, it is essential to note that exceedingly high values may lead to a decline in performance, given that the video does not exhibit complete consistency.

\subsection{Qualitative Analysis}
We provide a qualitative comparison between AVSBench~\cite{2023AVSS}, AVSegformer~\cite{2023avsegformer} and our proposed method on AVSBench-object and AVSBench-semantic datasets.
As depicted in Fig.~\ref{fig:figure_visualize}, our method, COMBO, exhibits superior audio-temporal and spatial localization quality, leading to better visualization and segmentation performance. 
For instance, in the case of the middle samples, our model accurately segments the singing man despite the presence of other sounds.
Moreover, our method achieves more precise segmentation for background noise handling and provides richer details of the foreground in other examples.

\section{Conclusion}
We introduce a novel audio-visual transformer framework, termed COMBO, that archives state-of-the-art performance on AVSBench-object and AVSBench-semantic datasets.
Contrary to previous methodologies that only factor in modality or temporal relations individually, our method explores multi-order bilateral relations for the first time, combining pixel entanglement, modality entanglement, and temporal entanglement.
For these three kinds of entanglement, we propose Siam-Encoder Module~(SEM), Bilateral-Fusion Module~(BFM), and adaptive inter-frame consistency loss, respectively.
Extensive experimental results verify the effectiveness of our proposed framework. 
We hope that our work will inspire further research in this significant and worthwhile field.

\vspace{-0.07cm}
\subsection*{Acknowledgements}
\vspace{-0.07cm}
\noindent This research was supported by the Strategic Priority Research Program of Chinese Academy of Sciences (Grant No.~XDB0500103), the National Natural Science Foundations of China (Grants No.~62076242, 62376267), the Pre-Research Project on Civil Aerospace Technologies (No.~D030312), the National Defense Basic Scientific Research Program of China~(No.~JCKY2021203B063) and the innoHK project.

\newpage
{
    \small
    \bibliographystyle{ieeenat_fullname}
    \bibliography{main}
}

\clearpage
\setcounter{page}{0}
\setcounter{section}{0}
\setcounter{figure}{0}
\setcounter{table}{0}
\renewcommand{\thesection}{\Alph{section}} 
\renewcommand{\thefigure}{\Roman{figure}} 
\renewcommand{\thetable}{\Roman{table}} 
\maketitlesupplementary

This appendix presents additional materials and results.

First, we describe the detailed training settings in Sec.~\ref{sec:supp_settings}.
Then, we give further descriptions of our proposed COMBO in Sec.~\ref{sec:supp_task_description} to enhance comprehension. 
Next, we provide more ablation studies for COMBO in Sec.~\ref{sec:supp_ablation_results}. 
Finally, a series of visual results are presented in Sec.~\ref{sec:supp_qualitive_results}.

\section{More Implementation Details}
\label{sec:supp_settings}
This section further explains the experimental details, which can be found in Tab.~\ref{tab:supp_settings}. 
It should be noted that the batch size pertains to the number of videos entered, thereby implying $bs \times T $ frames per iteration, where $bs$ denotes the batch size. 
Furthermore, pertaining to the AVSS task, given that the input video comprises varying numbers of frames, the number of frames within the batch size was dynamically altered without the need for padding zeros.

\section{Further Descriptions}  \label{sec:supp_task_description}
\subsection{Task Description}

\begin{figure}[t!]
    \centering
    \includegraphics[width=0.95\columnwidth]{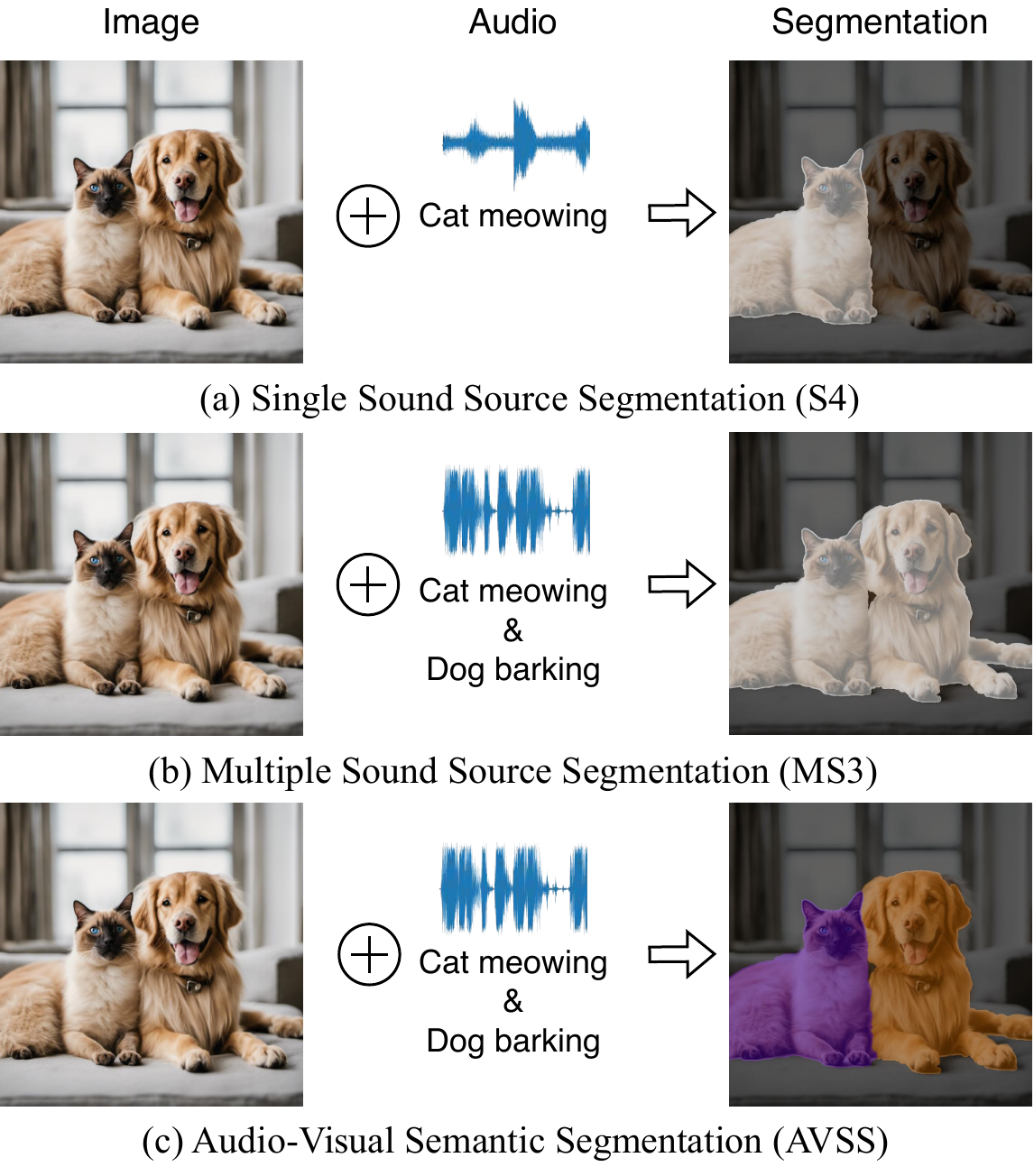}
    \caption{
    Illustration of the three sub-tasks in AVSBench-object and AVSBench-semantic datasets. 
    Best viewed in color.
    } 
    \label{fig:supp_figure1_task}
\end{figure}

We begin by providing an illustration description for the Audio-Visual Segmentation~(AVS). 
As depicted in Fig.~\ref{fig:supp_figure1_task}, the purpose of AVS is to segment all sound objects pixel-by-pixel. 
There are two datasets included: (1) \emph{AVSBench-object}. 
This dataset encompasses single source sound segmentation (S4) and multiple sound sources segmentation (MS3), as shown in Fig.~\ref{fig:supp_figure1_task}~(a) and Fig.~\ref{fig:supp_figure1_task}~(b), respectively.
In other words, objects (such as a dog or cat) in an image can be categorized into class-agnostic masks based on their corresponding sounds.
(2) \emph{AVSBench-semantic}. 
In addition to the above, objects that emit sounds also carry class semantic information, a concept known as audio-visual semantic segmentation (AVSS), as shown in Fig.~\ref{fig:supp_figure1_task}~(c).
This represents a more challenging dataset due to its complexity.


\begin{table*}[t!]
    \centering
    \small
    \vspace{-0.2cm}
    \setlength{\tabcolsep}{5pt} 
    \renewcommand{\arraystretch}{0.9} 
    \begin{tabular}{lccc} 
        \toprule 
         \textbf{Settings} & \textbf{S4}  & \textbf{MS3} & \textbf{AVSS} \\
        \midrule
         Resolution  $H \times W$   & $224 \times 224$ & $224 \times 224$ & $224 \times 224$ \\
         Number of frames $T$  &  5 & 5 & 5 \& 10  \\ 
         Data augmentation & horizontal flip \& color aug & horizontal flip \& color aug & horizontal flip \& color aug \\
         Audio dimension $D$ & 128 & 128 & 128 \\
         Embedding dimension $d$ & 256 &  256 & 256 \\
         Number of queries $N_q$ & 100 & 100 & 100 \\
         Number of transformer decoders $L$ & 3 & 3 & 3 \\
         Loss coefficient $\lambda_{cls}$ & 2.0 & 2.0 & 2.0 \\
         Loss coefficient $\lambda_{mask}$ & 5.0 & 5.0 & 5.0 \\
         Loss coefficient $\lambda_{ada}$ & 10.0 & 10.0 & 5.0 \\
         \midrule
         Batch size & 8 & 8 & 8 \\
         Optimizer & AdamW & AdamW & AdamW \\
         Learning rate & 0.0001 & 0.0001 & 0.0001 \\
         Weight decay & 0.05 & 0.05 & 0.05 \\
         Iterations & 90k & 20k & 90k \\
         
        \bottomrule 
    \end{tabular}
    \vspace{-0.2cm}
    \caption{Detailed settings. 
    This table provides a detailed overview of the specific settings used for each sub-task.}
    \label{tab:supp_settings}
\end{table*}

\begin{figure}[t!]
    \centering
    \includegraphics[width=1.0\columnwidth]{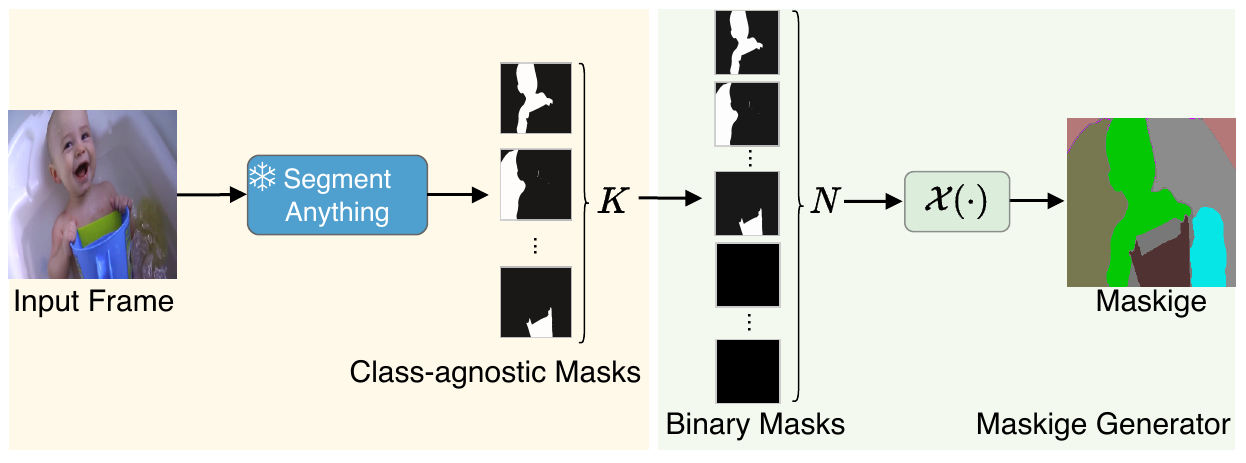}
    \caption{
    Illustration of the Proposal Generator. The proposal generator consists of two parts: the yellow area on the left mainly contains a frozen foundation model for generating class-agnostic masks, and the green area on the right is used to convert the masks into Maskige, also called Maskige generator.
    } 
    \label{fig:supp_figure3_maskige}
\end{figure}

\begin{figure*}[t!]
    \centering
    \includegraphics[width=1.8\columnwidth]{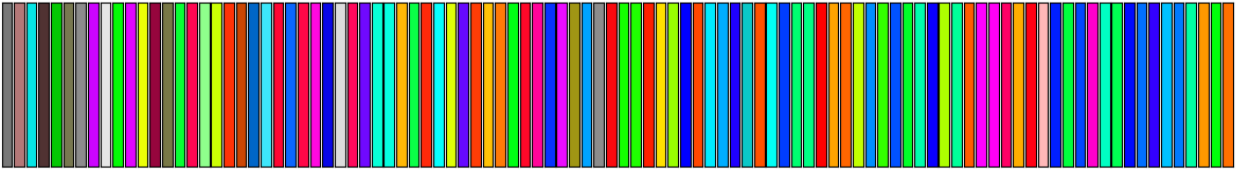}
    \caption{
    Visualization of matrix $ A \in \mathbb{R}^{100 \times 3}$. Each color bar has an RGB value of dimension 3, and there are 100 color bars.
    }
    \label{fig:supp_figure4_ade}
\end{figure*}

\subsection{Proposal Generator}
As shown in Fig.~\ref{fig:supp_figure3_maskige}, we provide a more detailed explanation of the proposal generator proposed in COMBO.
Initially, we obtain class-agnostic masks denoted as $c \in \mathbb{R}^{K \times H \times W}$ from the input frame, using a pre-existing foundation model~\cite{2023_semanticsam}, where $K$ denotes the number of potential targets.
Subsequently, a Maskige generator, which is part of the proposal generator, is introduced to convert the class-agnostic masks $c \in \mathbb{R}^{K \times H \times W}$ into Maskige, denoted as $m \in \mathbb{R}^{3 \times H \times W}$ without the need for additional training.
Particularly, as $K$ is dynamic and fluctuates according to input frames, we amplify the quantity of class-agnostic masks to $N$ using zero masks, thereby deriving a series of binary masks $c \in \mathbb{R}^{N \times H \times W}$, where $N$ is a predetermined number and $N \geq K$.
Next, considering that $c$ denotes a series of binary masks that are challenging to incorporate into visual features, we utilize a random color encoding function $\mathcal{X}(\cdot): \mathbb{R}^{N \times H \times W} \rightarrow \mathbb{R}^{3 \times H \times W} $ to convert the binary masks $c \in \mathbb{R}^{N \times H \times W}$ into Maskige $m \in \mathbb{R}^{3 \times H \times W}$.
Formally, we define $\mathcal{X}(c) = cA $, where $A \in \mathbb{R}^{N \times 3}$. 
To facilitate the proposal generator offline, the value of $A$ is manually set appropriately.
Specifically, we set $N=100$, a value considerably larger than $K$.
To enhance the distinctiveness between various targets, as illustrated in Fig.~\ref{fig:supp_figure4_ade}, we use the first 100 color mappings of the color mapping relationship in ADE20K dataset~\cite{2019ade_20k} as the parameters of matrix $A$.
Further visualizations on Maskiges are available in Sec.~\ref{sec:supp_qualitive_results}.

\begin{figure*}[t!]
    \centering
    \includegraphics[width=2.05\columnwidth]{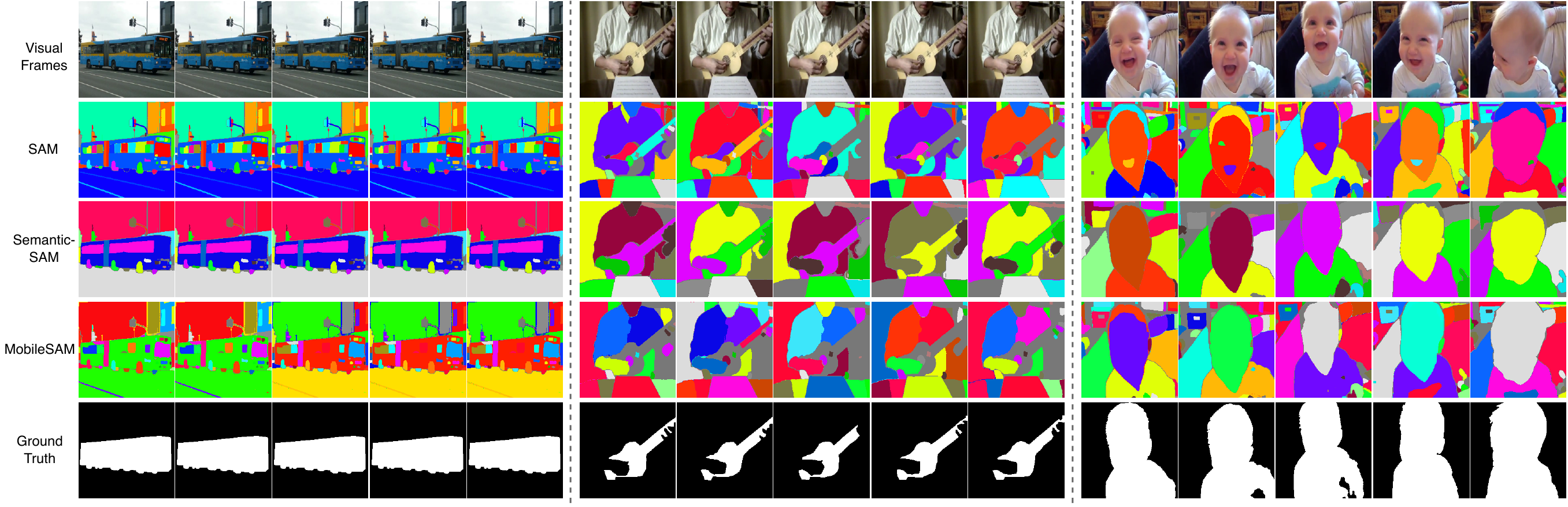}
    \caption{
    Visualization of Maskiges. The Maskiges are generated by proposal generator with various foundation models~\cite{2023_SAM_ICCV,2023_semanticsam,2023_mobile_sam} and the color encoding function $\mathcal{X}(\cdot)$.
    }
    \label{fig:supp_figure5_maskige_visual}
\end{figure*}

\section{More Results} 
\label{sec:supp_ablation_results}
\subsection{Effects of The Foundation Model}
We continue our exploration by investigating the impact of the various foundation models in the proposal generator on performance. 
For comparison, we select the original Segment Anything Model~(SAM)~\cite{2023_SAM_ICCV}, the superior-performing Semantic-SAM~\cite{2023_semanticsam}, and the lighter MobileSAM~\cite{2023_mobile_sam} as the foundation models of the proposal generator to evaluate the performance alongside the backbone of PVT-v2~\cite{2022_pvt} on the S4 subset.
As illustrated in Tab.~\ref{tab:supp_proposal}, the results depict a minimal performance discrepancy among the different foundational models.
Nevertheless, it is evident that the performance of our model improves with the enhancement of the foundational model's ability.
Accordingly, we choose the Semantic-SAM~\cite{2023_semanticsam} as the foundation model of the proposal generator. 
In addition, we also provide a comparison of the visualizations of the Maskiges generated by different foundational models in Sec.~\ref{sec:supp_qualitive_results}.

\subsection{Effects of The Number of Queries}
We present additional ablation studies concerning the number of queries, denoted as $N_q$, in our approach, as illustrated in Tab.~\ref{tab:supp_queries}.
In order to examine the influence of the query count on the model's performance, we conducted a series of experiments using varying quantities of queries within the transformer decoder, specifically 100, 200, and 300.
Our findings suggest that 100 queries are sufficient, given the infrequency of maximum concurrent classes in an AVS task.
Therefore, we established the default number of queries as 100 following~\cite{2022_Mask2Former}.

\section{More Qualitative Results }  \label{sec:supp_qualitive_results}

\begin{table}[t!]
    \centering
    \small
    \setlength{\tabcolsep}{4pt} 
    \renewcommand{\arraystretch}{1.0} 
    \begin{tabular}{lccc} 
        \toprule 
        \textbf{S4} & \textbf{SAM}~\cite{2023_SAM_ICCV} & \textbf{Semantic-SAM}~\cite{2023_semanticsam} & \textbf{MobileSAM}~\cite{2023_mobile_sam} \\
        \midrule
         $\mathcal{M_J}$ &  84.4 & \textbf{84.7} & 84.1  \\ 
         $\mathcal{M_F}$  & 91.8 & \textbf{91.9} &  91.6  \\ 
        \bottomrule 
    \end{tabular}

    \vspace{-0.2cm}
    \caption{Impact of the different foundation models on COMBO.}
    \label{tab:supp_proposal}
\end{table}
\begin{table}[t!]
    \centering
    \small
    \vspace{-0.2cm}
    \setlength{\tabcolsep}{13pt} 
    \renewcommand{\arraystretch}{1.0} 
    \begin{tabular}{ccccc} 
        \toprule 
         \multirow{2}*{$N_q$} & \multicolumn{2}{c}{\textbf{S4}}  & \multicolumn{2}{c}{\textbf{AVSS}} \\
        \cmidrule(lr){2-3}  \cmidrule(lr){4-5}  
        & $\mathcal{M_J}$ & $\mathcal{M_F}$ & $\mathcal{M_J}$ & $\mathcal{M_F}$  \\
        \midrule
         100 &  \textbf{81.7} & \textbf{90.1} & \textbf{33.3} & \textbf{37.3} \\ 
         200  & 81.5 & \textbf{90.1} & 32.0 & 35.6   \\   
         300  & 81.3  & 89.9 &  31.4 & 34.7   \\
        \bottomrule 
    \end{tabular}
    \vspace{-0.2cm}
    \caption{Impact of the number of queries on COMBO.}
    \vspace{-0.5cm}
    \label{tab:supp_queries}
\end{table}

\begin{figure}[b!]
    \centering
    \includegraphics[width=0.9\columnwidth]{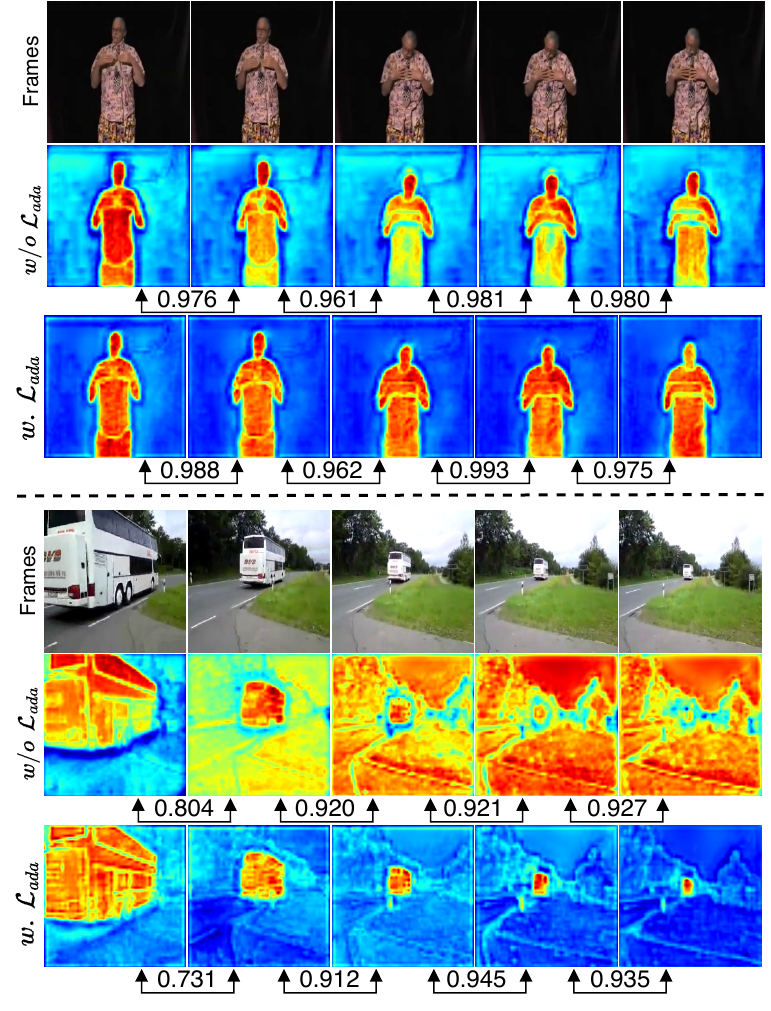}
    \caption{
    Visualization of the heat map of the predicted masks without and with the consideration of $\mathcal{L}_{ada}$ based on the S4 subset.
    } 
    \label{fig:supp_figure6_cosine}
    \vspace{-0.5cm}
\end{figure}

In this section, we introduce additional qualitative results of our proposed COMBO, along with its intermediate visualizations, to illustrate the effectiveness of our module.
The quality of the generation of the Maskige is crucial to the assistance of our model.
Therefore, we first show some examples sampled from
various sub-tasks with foundation models~\cite{2023_SAM_ICCV,2023_semanticsam,2023_mobile_sam} in Fig.~\ref{fig:supp_figure5_maskige_visual}.
It is evident that all foundational models exhibit exceptional proficiency in segmenting class-agnostic targets.
However, given that Semantic-SAM~\cite{2023_semanticsam} can produce a more complete target mask, we select it as the foundation model of our proposed proposal generator.
Besides, we also provide additional heat maps of the predicted masks to illustrate the effectiveness of the adaptive inter-frame consistency loss, $\mathcal{L}_{ada}$.
As depicted in Fig.~\ref{fig:supp_figure6_cosine}, when adjacent frames are similar, our loss module enables predicted masks to produce more accurate results.
Conversely, when adjacent frames are dissimilar, our module can avoid mutual interference between adjacent frames due to the existence of adaptive.
Finally, given that the audio-visual segmentation task is a video task with audio input, we present a visual comparison between our method and baseline in a video format, which can be reviewed on our project page.


\end{document}